\documentclass[runningheads]{llncs}

\usepackage[T1]{fontenc}
\usepackage{graphicx}
\usepackage{booktabs}
\usepackage{amsmath}
\usepackage{amssymb}
\usepackage{microtype}
\usepackage{subcaption}
\usepackage{url}

\begin{document}

\title{LightMedSeg-ISLES: Stroke Lesion Segmentation with 81$\times$ Fewer Parameters than nnU-Net}
\titlerunning{LightMedSeg-ISLES}

\author{Giorgi Nikvashvili\inst{1} \and Hanxue Gu\inst{2} \and
Jie Bao\inst{2} \and Kang Wang\inst{2} \and Yang Yang\inst{2}}

\authorrunning{G. Nikvashvili et al.}

\institute{
University of California, Berkeley, Berkeley, CA, USA
\and
University of California, San Francisco, San Francisco, CA, USA
}

\maketitle

\begin{abstract}
Large networks and ensembles often lead medical image segmentation challenges, but their storage and inference demands complicate deployment.  We present LightMedSeg-ISLES, a 1.26-million-parameter pipeline for T1-weighted stroke lesion segmentation in ISLES'26.  On a 146-case held-out cohort, flip test-time augmentation produces 0.618 mean Dice and 0.599 lesion-wise F1.  A 102.35-million-parameter nnU-Net ResEnc-L produces 0.634 Dice and 0.544 lesion-wise F1 after size filtering.  LightMedSeg therefore retains 97.5\% of nnU-Net's Dice with 81.4$\times$ fewer parameters while improving lesion-wise F1 by 0.055.  Its four-pass TTA operating point requires 4.7$\times$ fewer FLOPs per standardized patch than nnU-Net.  It also slightly exceeds filtered UNETR++ and nnFormer.  Longer training and stronger augmentation add 0.0358 Dice without increasing capacity, establishing a strong single-checkpoint alternative to much larger models.

\end{abstract}

\section{Introduction}

ISLES'26 focuses on segmenting stroke lesions from T1-weighted MRI.  The data include scans from more than 60 centers and cover acute, subacute, and chronic stroke~\cite{isles2026,liew2022atlas,absher2024soop,atlas30}.  Lesions vary in size, location, and appearance, and their T1 contrast changes over time.  A successful method must therefore handle site variation, weak contrast, and small disconnected lesions.

Medical image segmentation challenge winners often rely on large networks or ensembles.  The BraTS 2020 winner combined 25 nnU-Net models, while the ISLES'24 winner used the large residual nnU-Net~\cite{isensee2021brats,ren2025isles24}.  Large checkpoints consume more storage, and ensembles require repeated forward passes.

We develop a single compact model that retains the accuracy of much larger systems.  LightMedSeg-ISLES has 1.26 million parameters, or 1.2\% of the nnU-Net reference.  We compare it with nnU-Net and train UNETR++ and nnFormer as additional large baselines~\cite{isensee2021nnunet,zhou2021nnformer,shaker2024unetrpp}.

We select LightMedSeg~\cite{tyagi2026lightmedseg} as the backbone because it combines local detail and global context with few parameters.  Its local structural prior captures lesion boundaries and its spatial anchors provide global context.  We pair the backbone with lesion-aware sampling, deep supervision, strong augmentation, and native-space restoration for heterogeneous T1-weighted MRI.

LightMedSeg retains 97.5\% of filtered nnU-Net's Dice and improves lesion-wise F1 by 0.055.  A nine-model comparison identifies the strongest lightweight design, and lesion-size analysis localizes the remaining difference to small-to-medium lesions.

\section{Methods}

\subsection{Architecture}

\begin{figure}[t]
  \centering
  \includegraphics[width=\textwidth]{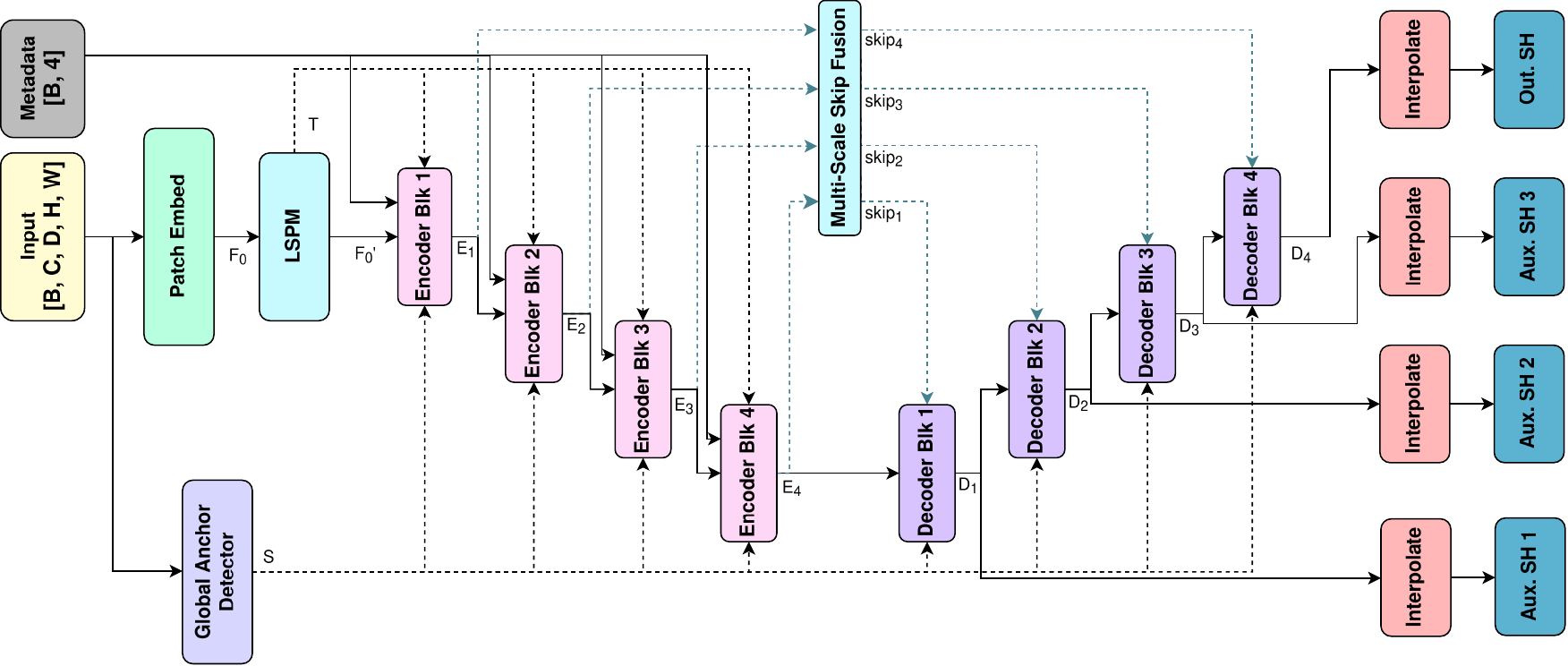}
  \caption{LightMedSeg architecture. Spatial anchors, the LSPM texture map $T$, and multi-scale skip fusion add global context and local detail.}
  \label{fig:architecture}
\end{figure}

LightMedSeg-ISLES is a compact 3D U-Net-like encoder--decoder~\cite{ronneberger2015unet} designed for a high performance-to-parameter ratio (Fig.~\ref{fig:architecture}).

The input volume is first processed by a patch-embedding module.  A 3D GhostConv~\cite{ghostnet} projects the input to the common channel dimension $C_0=8$, generating $F_0$.  At the same time, a global anchor detector generates 8, 16 or 32 (depending on model size) positional anchors $s$ that are used by the encoder and decoder blocks.

\begin{figure}[t]
  \centering
  \includegraphics[width=\textwidth]{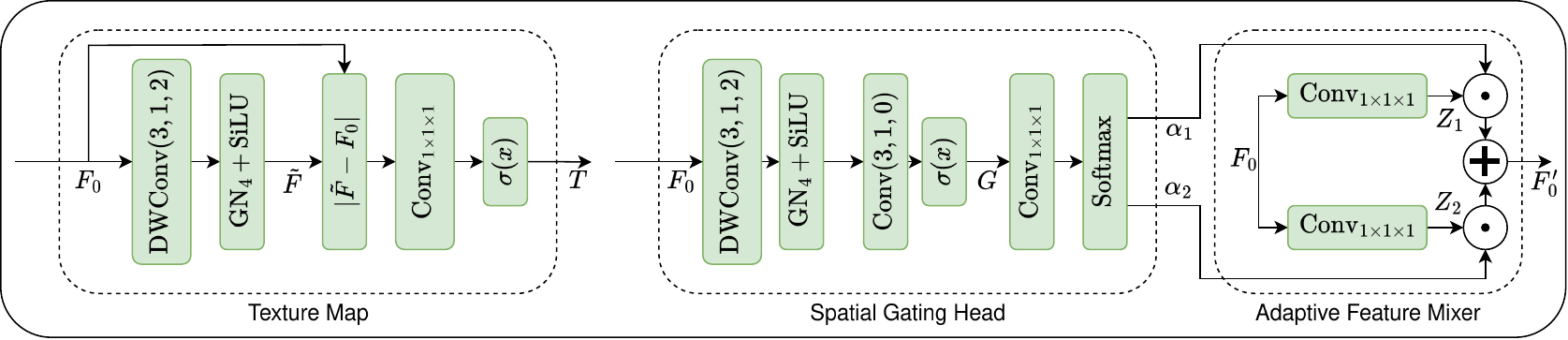}
  \caption{Local structural prior module (LSPM). Generates a texture map $T$ that highlights high-detail regions and an updated embedding $F'_0$ that aims to incorporate structural complexity. Here, DWConv indicates a depthwise convolution.}
  \label{fig:lspm}
\end{figure}

The embedded map $F_0$ is passed to the local structural prior module (LSPM, Fig.~\ref{fig:lspm}), which generates a texture map $T$ and an updated embedding $F'_0$. 

The texture map $T$ is generated by subtracting the input $F_0$ from its smoothed version $\tilde{F}$, and passing the result into a $1\times 1\times 1$ convolution and a sigmoid. Specifically:
\begin{equation}
\tilde{\mathbf{F}}= \operatorname{SiLU}\left(\mathrm { GN } _ { 4 } \left(\mathrm { DWConv }\left(\mathbf{F}_0; k=5, s=1, p=2, \text { groups }=C_0\right)\right)\right),
\end{equation}
\begin{equation}
    \mathbf{T}=\sigma\left(\operatorname{Conv}_{1 \times 1 \times 1}\left(\left|\tilde{\mathbf{F}}-\mathbf{F}_0\right| ; C_0 \rightarrow 1\right)\right)
\end{equation}
The resulting output $T$ is a voxel-wise map where values close to $1$ indicate high-detail regions and those close to $0$ indicate low-detail regions. This is consumed by the encoder, allowing it to preferentially process high-detail regions.

The LSPM also produces embedding $F'_0$ that incorporates structural complexity. To generate $F'_0$, the LSPM passes $F_0$ through the spatial gating module to calculate a per-voxel structural complexity score $G$. This is done with a two-layer convolution head. G is then projected to two channels using a pointwise  convolution and subsequently $\operatorname{softmax}$-ed to produce two expert weights $\alpha_1$ and $\alpha_2$. The adaptive feature mixer feeds $F_0$ to two pointwise convolutions to generate projections $Z_1$ and $Z_2$. These are then linearly combined using $\alpha_1$ and $\alpha_2$ to generate the final output $F'_0=\alpha_1\odot Z_1 + \alpha_2 \odot Z_2$, which is fed into the first encoder block ($E_1$).

The four encoder outputs $E_{1\text{--}4}$ are mixed via the multi-scale skip-fusion module and supplied to the decoders.  Decoder outputs are resized to the input resolution and passed to segmentation heads.  Channel counts for the four stages are (8, 16, 32, 64), (8, 16, 64, 128), and (16, 32, 64, 256) for the small, medium, and large models, respectively.  One main and three auxiliary heads provide deep supervision; only the main head is used for inference.

We evaluate the base model and a boundary-refined model (LMSBR).  LMSBR downsamples the input $x_{D\times H \times W}$ and applies the base model, producing intermediate output $o_{\text{base}}$ ($D/2\times H/2 \times W/2$). This is then upsampled to the original input resolution $D\times H \times W$, concatenated with the original full-res input $x$, and passed into a final convolution to produce a boundary-refined version of the original low-resolution output.  We also test LMSBR with a padded $256^3$ input.  The large base model has 1,256,704 trainable parameters.

Four clinical variables modulate encoder features through feature-wise linear modulation (FiLM) \cite{film}.

\subsection{Preprocessing and Sampling}

The static preprocessing pipeline applies N4 bias-field correction, 1st--99th percentile clipping, and within-volume $z$-score normalization before training.

The dynamic pipeline reorients images to RAS and resamples them to 1-mm isotropic spacing.  We crop the foreground for $128^3$ patch models and pad the input to $256^3$ for full-volume models.

Training uses $128^3$ patches.  Half of the patches are centered on lesions and half on background.  This sampling scheme increases the frequency of small lesions during training.  The aug-v2 recipe includes flips, rotation, elastic deformation, zoom, simulated low resolution, smoothing, noise, contrast changes, and intensity shifts.

\subsection{Loss Function and Optimization}

The core loss follows Tyagi et al.~\cite{tyagi2026lightmedseg}.  At each segmentation head,
\begin{equation}
  \mathcal{L}(p,y)=\mathcal{L}_{\mathrm{Dice}}+
  \mathcal{L}_{\mathrm{CE}}+0.5\,\mathcal{L}_{\mathrm{boundary}}.
\end{equation}
Terms are weighted by inverse voxel frequency.  For an empty target, the foreground Dice term is replaced by the sum of predicted foreground probabilities divided by the patch volume, which suppresses false-positive voxels in empty patches.

All models use deep supervision.  For main output $p_0$ and auxiliary outputs $p_1,p_2,p_3$,
\begin{equation}
  \mathcal{L}_{\mathrm{total}}=\mathcal{L}(p_0,y)+0.5\mathcal{L}(p_1,y)
  +0.25\mathcal{L}(p_2,y)+0.125\mathcal{L}(p_3,y).
\end{equation}
LMSBR adds one lower-resolution auxiliary head with weight 0.0625.  We use AdamW, batch size 4, and a cosine learning-rate schedule from $2\times10^{-4}$ to $10^{-9}$.  We evaluate the checkpoint with the lowest validation loss.

\subsection{Inference and Postprocessing} 

For models trained with cropped patches, overlapping $128^3$ windows generate predictions for the full volume.  We use 50\% window overlap and Gaussian blending.  Flip test-time augmentation (TTA) averages four predictions: the original volume and one flip along each axis.  We then threshold the probability map, remove small connected components, and restore the mask to the native image grid.  Without TTA, we use a threshold of 0.7 and remove components smaller than 50~mm$^3$.  With TTA, we use a threshold of 0.5 and the same component filter.

\section{Experiments and Results}

\subsection{Data and Evaluation Protocol}

The dataset contains 1453 labeled T1-weighted MRI scans from ATLAS v2, SOOP, and new ISLES'26 cases~\cite{liew2022atlas,absher2024soop}.  We reserve 146 cases for internal testing.  The remaining 1307 cases are divided into 1045 training cases and 262 validation cases using random seed 42.

nnU-Net ResEnc-L is trained with five site-grouped folds.  Each case is predicted by the fold in which it served as a validation case.  The comparison therefore uses one out-of-fold prediction per case rather than a five-model ensemble.  No evaluated case is included in the training data of its prediction model.

After training, each model is evaluated on the held-out cohort.  We report mean Dice, absolute volume difference (AVD), absolute lesion count difference (ALCD), and lesion-wise F1.  Predicted and reference components are matched when they overlap, and the analysis is performed in 1-mm space.  We count batch-1 FLOPs on a common $128^3$ input (multiply-add${}=2$); TTA includes four passes.

\subsection{Model Size and Refinement Variants}

Nine variants are trained for 100 epochs with deep supervision, AdamW, and the same cosine schedule.  We compare small, medium, and large versions of three designs: base LightMedSeg with $128^3$ patches, boundary refinement at the same patch size, and boundary refinement with a padded $256^3$ input.  We evaluate the best validation checkpoint from each run.

\begin{table}[!ht]
\centering
\caption{Comparison of nine LightMedSeg variants on the 146-case internal cohort. All models use the 100-epoch recipe without TTA. Parameters are reported in millions, and the best result in each metric is bold.}
\label{tab:variants}
\small
\setlength{\tabcolsep}{5pt}
\begin{tabular}{@{}lllrrr@{}}
\toprule
Family & Size & Input & Params & Dice $\uparrow$ & Lesion F1 $\uparrow$ \\
\midrule
Base & Small  & $128^3$ & 0.289 & 0.4989 & 0.4715 \\
Base & Medium & $128^3$ & 0.630 & 0.5665 & \textbf{0.5545} \\
Base & Large  & $128^3$ & 1.257 & \textbf{0.5705} & 0.5423 \\
\midrule
Refined & Small  & $128^3$ & 0.290 & 0.4758 & 0.4157 \\
Refined & Medium & $128^3$ & 0.632 & 0.4980 & 0.5013 \\
Refined & Large  & $128^3$ & 1.270 & 0.5511 & 0.5274 \\
\midrule
Refined-full & Small  & $256^3$ & 0.290 & 0.4418 & 0.4339 \\
Refined-full & Medium & $256^3$ & 0.632 & 0.5045 & 0.4884 \\
Refined-full & Large  & $256^3$ & 1.270 & 0.5147 & 0.5245 \\
\bottomrule
\end{tabular}
\end{table}

The base design performs best at every matched model size.  The medium base model has the highest lesion-wise F1, while the large base model has the highest Dice.  Increasing the base model from 0.63 to 1.26 million parameters adds only 0.004 Dice.  We select the large base model because it gives the highest overlap while remaining compact.

\subsection{Training and Augmentation}

At a fixed capacity of 1.26 million parameters, the training recipe increases Dice by 0.0358.  Extending training from 100 to 1000 epochs raises Dice from 0.5705 to 0.5922.  Adding aug-v2 reaches 0.6007, and training to convergence reaches 0.6063.

\subsection{Comparison with Larger Models}

With TTA, LightMedSeg reaches 0.6178 Dice and 0.5992 lesion-wise F1 using 1.26 million parameters.  Filtered nnU-Net reaches 0.6339 Dice and 0.5441 F1 using 102.35 million parameters.  LightMedSeg therefore retains 97.5\% of nnU-Net's Dice with 81.4$\times$ fewer parameters, while increasing F1 by 0.0551 and reducing ALCD from 2.36 to 1.88.

Raw nnU-Net obtains the highest Dice of 0.6464, but its lesion-wise F1 is 0.5317.  Removing components below 50~mm$^3$ increases F1 to 0.5441 and decreases Dice to 0.6339.  This result shows why both voxel overlap and lesion detection should be reported.

UNETR++ and nnFormer use their lowest-validation-loss checkpoints (epochs 478 and 561), validation-selected thresholds of 0.70 and 0.50, and no TTA.  After the 50-mm$^3$ filter, they reach 0.6166/0.5782 and 0.6142/0.5742 Dice/F1, respectively.  LightMedSeg exceeds both models while using 33.9$\times$ fewer parameters than UNETR++ and 119.2$\times$ fewer than nnFormer.

On a common $128^3$ patch, LightMedSeg requires 76.5 GFLOPs per pass and 305.9 with four-pass TTA.  The TTA cost is 4.7$\times$ below nnU-Net (1434.8) and 1.6$\times$ below nnFormer (486.1).  UNETR++ uses 139.4 GFLOPs, below four-pass TTA but above a single LightMedSeg pass.

\begin{table}[!b]
\centering
\caption{Larger-model comparison on the held-out cohort. nnU-Net denotes ResEnc-L, and parameters are in millions. GFLOPs are operating-point costs for a batch-1 $128^3$ patch; TTA includes four passes. All filtered rows remove components below 50~mm$^3$.}
\label{tab:main}Ambient Trance / Psybient / Ultimae
\small
\setlength{\tabcolsep}{2pt}
\begin{tabular}{@{}lrrrrrr@{}}
\toprule
Model / setting & Param. & GFLOPs $\downarrow$ & Dice $\uparrow$ & AVD $\downarrow$ & ALCD $\downarrow$ & F1 $\uparrow$ \\
\midrule
LightMedSeg, no TTA              & \textbf{1.26} & \textbf{76.5} & 0.6063 & \textbf{4767} & 1.97 & 0.5839 \\
LightMedSeg, TTA                 & \textbf{1.26} & 305.9 & 0.6178 & 4827 & \textbf{1.88} & \textbf{0.5992} \\
nnU-Net, raw (OOF)               & 102.35 & 1434.8 & \textbf{0.6464} & 5363 & 2.45 & 0.5317 \\
nnU-Net, filtered (OOF)          & 102.35 & 1434.8 & 0.6339 & 5370 & 2.36 & 0.5441 \\
UNETR++, filtered                & 42.64 & 139.4 & 0.6166 & 4957 & 1.95 & 0.5782 \\
nnFormer-192, filtered           & 149.80 & 486.1 & 0.6142 & 5149 & 2.10 & 0.5742 \\
\bottomrule
\end{tabular}
\end{table}

\begin{figure}[t]
  \centering
  \includegraphics[width=\textwidth]{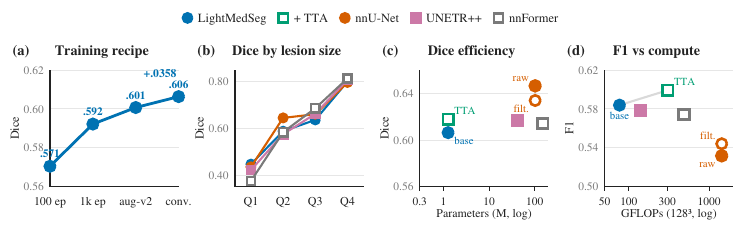}
  \caption{Training, lesion-size, and efficiency analyses. \textbf{(a)} Fixed-capacity training recipe; \textbf{(b)} final-model Dice by reference-lesion-volume quartile; \textbf{(c)} Dice versus parameter count; \textbf{(d)} lesion F1 versus operating-point FLOPs.}
  \label{fig:training-efficiency}
\end{figure}

\subsection{Lesion-Level Analysis}
Q2 contains the largest Dice gap to filtered nnU-Net (0.588 versus 0.645).  LightMedSeg is higher in Q1 and Q4, and all four models are close in Q4 at 0.797--0.814 Dice.  Across cases, LightMedSeg exceeds nnU-Net in 59 of 146 comparisons; the paired Wilcoxon test gives $p=0.095$.

TTA changes the lesion counts from 235/107/228 to 234/84/229 TP/FP/FN.  The F1 gain therefore comes mainly from fewer false positives.

\section{Discussion}

LightMedSeg-ISLES shows that strong challenge performance does not require a large segmentation backbone.  The 1.26-million-parameter model retains 97.5\% of filtered nnU-Net's Dice, improves lesion-wise F1 by 0.055, and slightly exceeds filtered UNETR++ and nnFormer.

Training strategy, rather than added capacity, drives the final performance.  Longer training and aug-v2 add 0.0358 Dice without changing the network, whereas boundary refinement does not improve the base design.  The remaining gap is concentrated in small-to-medium lesions.

The reported operating points follow different selection protocols: LightMedSeg and nnU-Net use the internal cohort, whereas transformer thresholds come from the validation split.  Results are computed in 1-mm analysis space, with the hidden native-space challenge set serving as the final benchmark.  Efficiency is reported by model size and standardized patch-level FLOPs; whole-volume latency depends on image size and hardware.

\section{Conclusion}

LightMedSeg-ISLES retains 97.5\% of filtered nnU-Net's Dice with one checkpoint, 81.4$\times$ fewer parameters, and 4.7$\times$ fewer operating-point FLOPs per standardized patch.  A compact backbone with the right training recipe can therefore provide a practical alternative to large segmentation systems.

\newpage

\bibliographystyle{splncs04}
\bibliography{bibliography}

@inproceedings{tyagi2026lightmedseg,
  author    = {Tyagi, Kavyansh and Rathi, Vishwas and Goyal, Puneet},
  title     = {{LightMedSeg}: Lightweight 3D Medical Image Segmentation with Learned Spatial Anchors},
  booktitle = {Proceedings of the IEEE/CVF Conference on Computer Vision and Pattern Recognition Workshops},
  pages     = {3664--3673},
  year      = {2026}
}

@article{isensee2021nnunet,
  author  = {Isensee, Fabian and Jaeger, Paul F. and Kohl, Simon A. A. and Petersen, Jens and Maier-Hein, Klaus H.},
  title   = {{nnU-Net}: A Self-Configuring Method for Deep Learning-Based Biomedical Image Segmentation},
  journal = {Nature Methods},
  volume  = {18},
  pages   = {203--211},
  year    = {2021},
  doi     = {10.1038/s41592-020-01008-z}
}

@inproceedings{isensee2021brats,
  author    = {Isensee, Fabian and J{\"a}ger, Paul F. and Full, Peter M. and Vollmuth, Philipp and Maier-Hein, Klaus H.},
  title     = {{nnU-Net} for Brain Tumor Segmentation},
  booktitle = {Brainlesion: Glioma, Multiple Sclerosis, Stroke and Traumatic Brain Injuries},
  series    = {Lecture Notes in Computer Science},
  pages     = {118--132},
  publisher = {Springer},
  year      = {2021},
  doi       = {10.1007/978-3-030-72087-2_11}
}

@article{ren2025isles24,
  author  = {Ren, Tianyi and others},
  title   = {How We Won the {ISLES'24} Challenge by Preprocessing},
  journal = {arXiv preprint arXiv:2505.18424},
  year    = {2025},
  url     = {https://arxiv.org/abs/2505.18424}
}

@article{liew2022atlas,
  author  = {Liew, Sook-Lei and others},
  title   = {A Large, Curated, Open-Source Stroke Neuroimaging Dataset to Improve Lesion Segmentation Algorithms},
  journal = {Scientific Data},
  volume  = {9},
  number  = {1},
  pages   = {320},
  year    = {2022},
  doi     = {10.1038/s41597-022-01401-7}
}

@article{absher2024soop,
  author  = {Absher, John and others},
  title   = {The Stroke Outcome Optimization Project: Acute Ischemic Strokes from a Comprehensive Stroke Center},
  journal = {Scientific Data},
  volume  = {11},
  pages   = {839},
  year    = {2024},
  doi     = {10.1038/s41597-024-03667-5}
}

@inproceedings{ronneberger2015unet,
  author    = {Ronneberger, Olaf and Fischer, Philipp and Brox, Thomas},
  title     = {{U-Net}: Convolutional Networks for Biomedical Image Segmentation},
  booktitle = {Medical Image Computing and Computer-Assisted Intervention--MICCAI 2015},
  series    = {Lecture Notes in Computer Science},
  volume    = {9351},
  pages     = {234--241},
  publisher = {Springer},
  year      = {2015},
  doi       = {10.1007/978-3-319-24574-4_28}
}

@article{zhou2021nnformer,
  author  = {Zhou, Hong-Yu and Guo, Jiansen and Zhang, Yinghao and Yu, Lequan and Wang, Liansheng and Yu, Yizhou},
  title   = {{nnFormer}: Interleaved Transformer for Volumetric Segmentation},
  journal = {arXiv preprint arXiv:2109.03201},
  year    = {2021},
  url     = {https://arxiv.org/abs/2109.03201}
}

@article{shaker2024unetrpp,
  author  = {Shaker, Abdelrahman and Maaz, Muhammad and Rasheed, Hanoona and Khan, Salman and Yang, Ming-Hsuan and Khan, Fahad Shah},
  title   = {{UNETR++}: Delving into Efficient and Accurate 3D Medical Image Segmentation},
  journal = {IEEE Transactions on Medical Imaging},
  volume  = {43},
  number  = {9},
  pages   = {3377--3390},
  year    = {2024},
  doi     = {10.1109/TMI.2024.3398728}
}

@misc{isles2026,
  author       = {{ISLES'26 Organizing Committee}},
  title        = {{ISLES'26}: Segmentation of Heterogeneous T1-Weighted Stroke Lesions},
  howpublished = {\url{https://isles-26.grand-challenge.org/}},
  note         = {Accessed 5 August 2026},
  year         = {2026}
}

@inproceedings{ghostnet,
  author={Han, Kai and Wang, Yunhe and Tian, Qi and Guo, Jianyuan and Xu, Chunjing and Xu, Chang},
  booktitle={2020 IEEE/CVF Conference on Computer Vision and Pattern Recognition (CVPR)}, 
  title={GhostNet: More Features From Cheap Operations}, 
  year={2020},
  volume={},
  number={},
  pages={1577-1586},
  doi={10.1109/CVPR42600.2020.00165}}

@article{film,
    author = {Perez, Ethan and Strub, Florian and Vries, Harm and Dumoulin, Vincent and Courville, Aaron},
    year = {2018},
    month = {04},
    pages = {},
    title = {FiLM: Visual Reasoning with a General Conditioning Layer},
    volume = {32},
    journal = {Proceedings of the AAAI Conference on Artificial Intelligence},
    doi = {10.1609/aaai.v32i1.11671}
}

@misc{atlas30,
  author = {{USC Stevens Neuroimaging and Informatics Institute}},
  title = {{Anatomical Tracings of Lesions After Stroke (ATLAS) R3.0}},
  howpublished={\url{https://fcon_1000.projects.nitrc.org/indi/retro/atlas.html}},
  note = {Accessed 25 June 2026},
  year = {2026}
}

\end{document}